\documentclass[10pt,twocolumn,letterpaper]{article}

\usepackage[algorithms]{wacv}   

\definecolor{wacvblue}{rgb}{0.21,0.49,0.74}
\usepackage[pagebackref,breaklinks,colorlinks,allcolors=wacvblue]{hyperref}
\usepackage{paralist}
\usepackage[inline]{enumitem}
\usepackage{enumitem}
\setlist[itemize]{noitemsep,topsep=0pt,leftmargin=*} 
\usepackage{adjustbox}
\usepackage{comment}
\usepackage{booktabs}   
\usepackage{multirow}   
\usepackage{graphicx}   
\usepackage{ulem}
\usepackage{dblfloatfix}
\usepackage{hyphenat}
\usepackage{mathtools}
\usepackage{siunitx}
\usepackage{pifont}

\def\wacvPaperID{2360} 
\def\confName{WACV}
\def\confYear{2026}

\newcommand{\model}{\textsc{TiCLS}}

\title{ \model: Tightly Coupled Language Text Spotter }

\author{
Leeje Jang$^{1}$ \quad Yijun Lin$^{1}$ \quad Yao-Yi Chiang$^{1}$ \quad Jerod Weinman$^{2}$ \\
$^{1}$University of Minnesota, United States \\
$^{2}$Grinnell College, United States \\
{\tt\small \{jang0124, lin00786, yaoyi\}@umn.edu \quad jerod@acm.org}
}

\begin{document}
\maketitle
\begin{abstract}
Scene text spotting aims to detect and recognize text in real-world images, where instances are often short, fragmented, or visually ambiguous. Existing methods primarily rely on visual cues and implicitly capture local character dependencies, but they overlook the benefits of external linguistic knowledge. 
Prior attempts to integrate language models either adapt language modeling objectives without external knowledge or apply pretrained models that are misaligned with the word-level granularity of scene text.
We propose \model, an end-to-end text spotter that explicitly incorporates external linguistic knowledge from a character-level pretrained language model.
\model\ contains a pretrained linguistic decoder that fuses visual and linguistic features, enabling robust recognition of ambiguous or fragmented text.
Experiments on ICDAR 2015, Total-Text, and CTW1500 demonstrate that \model\ achieves state-of-the-art performance, validating the effectiveness of PLM-guided linguistic integration for scene text spotting. The code is available at \url{https://github.com/knowledge-computing/TiCLS}. 

\end{abstract}

\section{Introduction}

\setlength{\unitlength}{1in}
\begin{picture}(0,0)\put(0,7.1){
\parbox[t]{6.8in}{\footnotesize
Copyright \copyright 2026 IEEE.
Reprinted from B. Clipp, N. Damer, S. X. Huang and V. Morariu, editors,
\emph{Proc. IEEE/CVF Winter Conference on Applications of Computer Vision (WACV)}, March 2026.
}}
\end{picture}



End-to-end scene text spotting detects and recognizes text from scene images in a unified framework. 
The visual appearance of scene text varies widely, often featuring unusual fonts, perspective distortion, unconventional typography, and even occlusions~\cite{tom2024occluded}. Nevertheless, linguistic patterns remain present in the underlying text.
Existing text spotters typically rely on convolution- or transformer-based architectures to extract visual features and use these features to predict text. 
However, text in scene images often appears degraded (blurred, noisy, or occluded), making it insufficient to rely solely on visual cues.

To address this limitation, some prior work incorporates linguistic context into text spotting.
For instance, LMTextSpotter~\cite{xia2024lmtextspotter} incorporates a parallel recognition branch explicitly trained for autoregressive language modeling. 
The language modeling contextualizes each character representation using the representations of surrounding characters within the same word during training. 
However, this contextualization relies solely on the text available in scene text spotting datasets, since the model lacks the ability to pretrain on large-scale text corpora.
As a result, the model fails to exploit abundant unlabeled text data and learns contextual patterns only from the limited training text corpora.

Pretrained language models (PLMs)~\cite{devlin2019bert, liu2019roberta, radford2019language, lewis2019bartdenoisingsequencetosequencepretraining, raffel2020exploring}, trained on large-scale corpora such as news articles and books, provide external linguistic knowledge by capturing semantic dependencies across sub-word tokens in well-structured sentences. 
TextBlockV2~\cite{lyu2025textblockv2} incorporates such external linguistic knowledge by detecting coarse, multi-word text blocks and then applying a PLM-based recognizer to transcribe the content within each text block instead of individual words.
This approach initializes the recognizer with an existing PLM, which autoregressively decodes text sequences for each block, thereby producing multi-phrase outputs when a block contains multiple lines or phrases. 
This method highlights the potential of leveraging PLMs in text spotting, especially for images containing large chunks of text, but challenges remain due to the short and fragmented nature of scene text.

Scene text differs significantly from well-structured, sentence-level language: instances are short, spatially fragmented, and often lack coherent grammatical structure. 
Simply initializing a recognition decoder with existing PLMs is therefore insufficient, as this initialization does not address the mismatch between sentence-level linguistic modeling and the word-level semantics required for scene text spotting. 
The key challenge is how to effectively adapt PLMs’ linguistic knowledge to scene text spotting so that linguistic cues serve as complementary information to support ambiguous or fragmented visual information. 

To bridge this gap, we introduce \model\ (\underline{Ti}ghtly \underline{C}oupled \underline{L}anguage \underline{S}potter), a novel end-to-end text spotting model that integrates external linguistic knowledge from a pretrained language model (PLM) specifically developed for scene text. \model\ consists of three main components: a visual feature extractor, a visual decoder, and a linguistic decoder. Following prior transformer-based text spotters~\cite{Huang_2023_ICCV,ye2023deepsolo,zhang2022text}, the visual feature extractor and decoder learn visual representations of scene text. 
To provide strong external linguistic knowledge tailored for scene text spotting, we propose a new transformer-based encoder-decoder PLM that we pretrain on a large text corpus with character-level prediction granularity instead of sub-word tokens.
We then initialize \model's linguistic decoder with this PLM, allowing the model to fuse our external linguistic priors with visual features in a tightly coupled manner to produce more accurate and robust text predictions.



We summarize our contributions as follows:
\begin{itemize}

\item We propose \model, a novel end-to-end text spotter that explicitly incorporates linguistically rich prior knowledge for scene text spotting.

\item We develop a character-level pretrained language model tailored for scene text, providing linguistic knowledge for short and fragmented text instances.

\item We introduce a linguistic decoder initialized with our PLM, enabling tight fusion of linguistic and visual information and improving recognition of ambiguous or visually degraded text.

\end{itemize}

    

\begin{figure*}[t]
  \centering \includegraphics[height=0.36\textheight,keepaspectratio]{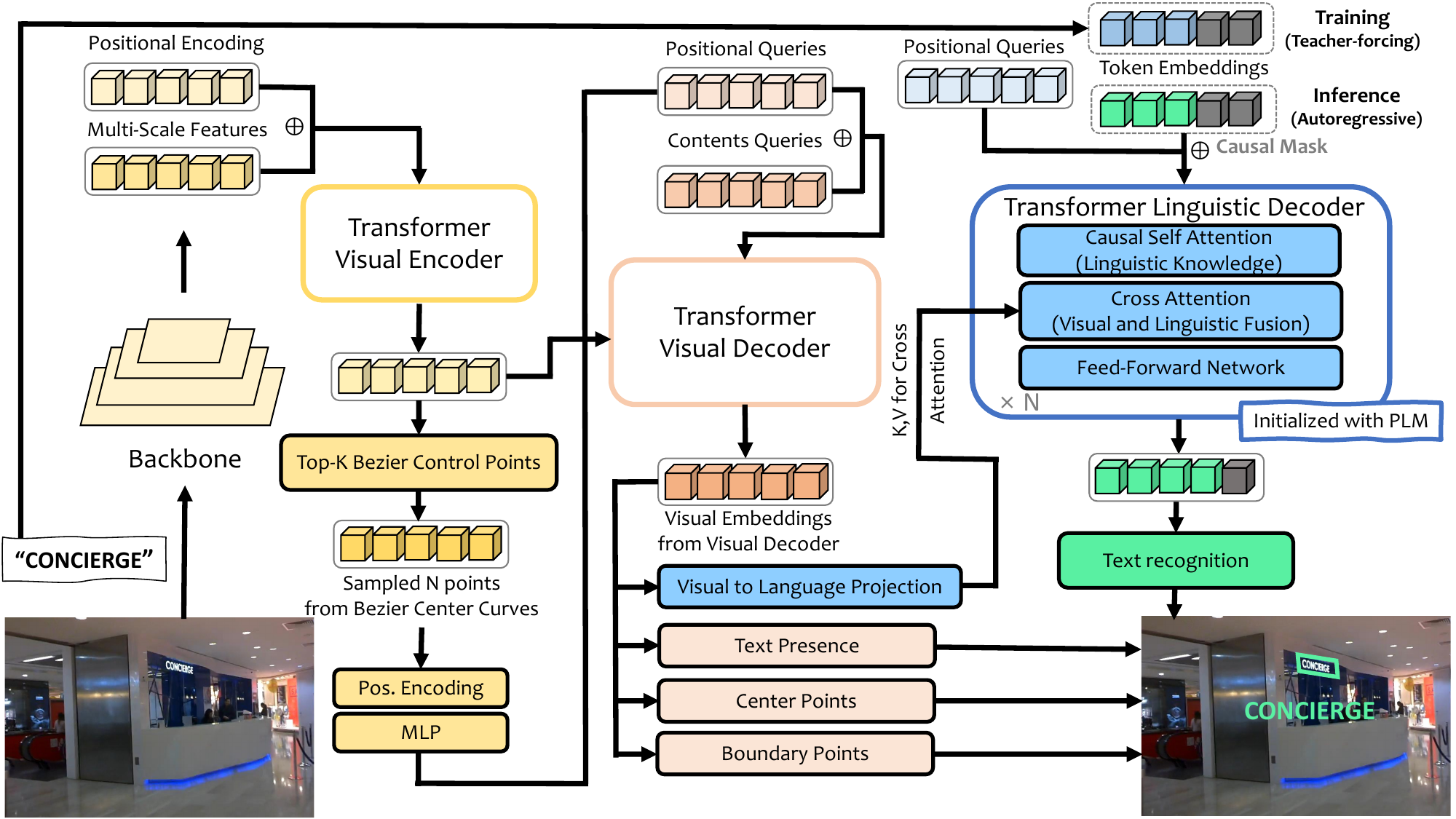
}
\caption{Overall architecture of \model. \model\ builds on a DETR-style visual encoder (yellow) and visual decoder (orange). A linguistic decoder (blue) receives visual representations from the visual-to-language projection head and performs visual-linguistic fusion for text recognition (green). The linguistic decoder is initialized with our proposed PLM (Section~\ref{subsec:plmtrain}).}
  \label{fig:overall_model}
\end{figure*}

\section{Related Work}\label{sec:relatedwork}
\subsection{End-to-end Scene Text Spotting}
End-to-end scene text spotting ~\cite{liu2018fots,zhang2022text,liu2020abcnet,ye2023deepsolo,Huang_2023_ICCV,liao2020mask,das2024fasttextspotter,huang2022swintextspotter,xia2024lmtextspotter} localizes and recognizes text from natural scene images within a single network.
These type of approaches enable shared feature representations between text detection and recognition, and jointly optimizes both tasks, leading to improved performance.

The increasing success of transformer models has expanded their application to computer vision tasks. DETR-based text spotting methods~\cite{ye2023deepsolo,huang2022swintextspotter,Huang_2023_ICCV,zhang2022text,xia2024lmtextspotter,lin2024hyper} have achieved state-of-the-art performance. 
TESTR~\cite{zhang2022text} consists of a single encoder and dual decoders, one for predicting control points localizing text regions and the other for character recognition.
DeepSolo~\cite{ye2023deepsolo} performs  transformer-based text spotting with a single decoder, localizing text regions with Bézier curves.
ESTextSpotter~\cite{Huang_2023_ICCV} enables explicit communication between the detection and recognition tasks. 
SwinTextSpotter v2~\cite{huang2025swintextspotter} proposes Recognition Conversion and Recognition Alignment modules to strengthen the synergy between detection and recognition. While these end-to-end text spotters demonstrate strong performance, the explicit use of prior linguistic knowledge has remained largely unexplored until recently. 

\subsection{Language-Aware Scene Text Spotting}\label{sub:language-spotting}
Recent work in scene text spotting has explored incorporating linguistic clues to enhance performance. LMTextSpotter~\cite{xia2024lmtextspotter} introduces a separate language modeling recognition branch trained in an autoregressive manner to capture character-level dependencies within a text sequence. This autoregressive design aligns with modeling strategies used in natural language processing (NLP), where decoder-only transformer models such as GPT~\cite{radford2019language} generate text by predicting the next token based on previously generated tokens.
Because LMTextSpotter trains only on text from scene images, it cannot fully benefit from the linguistic priors offered by broadly pretrained language models.

Existing pretrained language models (PLMs)~\cite{devlin2019bert, liu2019roberta, radford2019language, lewis2019bartdenoisingsequencetosequencepretraining, raffel2020exploring} are generally trained on large-scale text corpora such as news articles and books and are designed to learn the linguistic context of tokens within well-structured sentences. 
TextBlockV2~\cite{lyu2025textblockv2} leverages the pretrained GPT-2~\cite{radford2019language} to initialize the text spotting decoder. 
This approach extracts visual embeddings for each multi-word text block using an encoder, and then decodes these embeddings to produce block-level text recognition outputs.
This design attempts to leverage the ability of PLMs to capture linguistic information across sentence-level contexts.
However, because PLMs typically learn to capture complete sentences with coherent semantic meaning, they do not naturally align with scene text spotting, where text often appears as short, spatially localized words or phrases that rarely combine into a complete sentence.
As a result, such block-level approaches struggle to achieve consistently strong performance on standard scene text spotting benchmarks.

 \subsection{Text Recognition with Language Modeling}\label{subsec:str_lm}

Several works have incorporated language modeling into the scene text recognition (STR) task~\cite{wang2021from,bautista2022scene,li2023trocr}, which differs from text spotting in that it skips detection and assumes cropped text regions are directly provided as input.

Two general approaches exist that differ from \model\ in mechanism or decoding granularity. 
First, some approaches adopt language modeling objectives from NLP tasks while maintaining a character-level decoding space, including masked language modeling~\cite{wang2021from}.
For example, PARSeq~\cite{bautista2022scene} uses permutation language modeling~\cite{yang2019xlnet} for causal self-attention and cross-attends to visual encoder outputs with decoder-side linguistic embeddings. 
This approach helps STR to learn linguistic regularities implicitly.
TrOCR~\cite{li2023trocr} directly integrates a pretrained language model (PLM) into the text recognition task, decoding over sub-word tokens.
As described in Section~\ref{sub:language-spotting}, integrating existing PLMs for text recognition tends to perform well on long text sequences, such as lines, phrases, or sentences (e.g., handwritten text recognition), rather than on word-level text from STR.
Although recent STR work incorporates linguistic cues with notable success, integrating linguistic information into end-to-end scene text spotting remains underexplored, which forms a core contribution of this work.

\section{\model\ for Text Spotting} \label{sec:method}
\model\ consists of three components (Figure~\ref{fig:overall_model}): a visual feature extractor, a visual decoder, and a linguistic decoder. 
For the visual components, we adapt DeepSolo~\cite{ye2023deepsolo}, a DETR-based end-to-end text spotter.
We remove DeepSolo's character classification head and introduce a new visual-to-language projection module to map the visual features into the input space required by the linguistic decoder.
The linguistic decoder fuses the transformed visual features with prior knowledge by initializing the decoder with a pretrained language model specifically designed for scene text.

Section~\ref{subsec:visual} reviews the visual feature extractor and decoder; 
Section~\ref{subsec:plmtrain} introduces the overall architecture and the pretraining of our proposed language model; 
and Section~\ref{subsec:lingdec} describes the linguistic decoder in \model, which fine-tunes the integrated PLM decoder.


\subsection{Visual Feature Extractor and Decoder}\label{subsec:visual}

\model\ builds on DeepSolo's~\cite{ye2023deepsolo} visual feature extractor and decoder.
The visual feature extractor leverages a ViTAEv2-based backbone to obtain multi-scale visual features, followed by a visual encoder.
The visual encoder produces text region proposals together with their text-existence scores, and \model\ selects the top $K$ proposals based on these scores for further processing.
For each proposal, we sample $N$ points along the Bézier curves derived from the encoder output.
We apply a sinusoidal positional encoding function to these points and feed the encoded representations into an MLP to obtain the position queries for the visual decoder.
In the visual decoder, we add these position queries to the learnable content queries, and the resulting representations are iteratively refined through intra- and inter-sequence self-attention, which captures relationships among character-level visual features, and through cross-attention with the visual encoder outputs.

The decoder produces \(H^{\mathrm{vis}}\!\in\!\mathbb{R}^{K\times N\times D_\mathrm{V}}\) as the input to three heads:
\begin{inparaenum}[i)]
\item a binary text-presence classifier, 
\item a center points head regressing centerline Bézier offsets, and 
\item a boundary points head regressing top/bottom Bézier offsets.\end{inparaenum}

To bridge the visual and linguistic decoders, we introduce a visual-to-language projection head that maps embeddings from the visual decoder's ${D_\mathrm{V}}$-dimensional space into the linguistic decoder's ${D_\mathrm{L}}$-dimensional space.
The projected embeddings serve as the input to the linguistic decoder (Section~\ref{subsec:lingdec}), which performs text recognition with tightly coupled visual and linguistic representations.

\subsection{Pretrained Language Model for Scene Text }\label{subsec:plmtrain}
\begin{figure}[t] 
  \centering
    \includegraphics[
    width=\linewidth,
    height=2\textheight,
    keepaspectratio
  ]{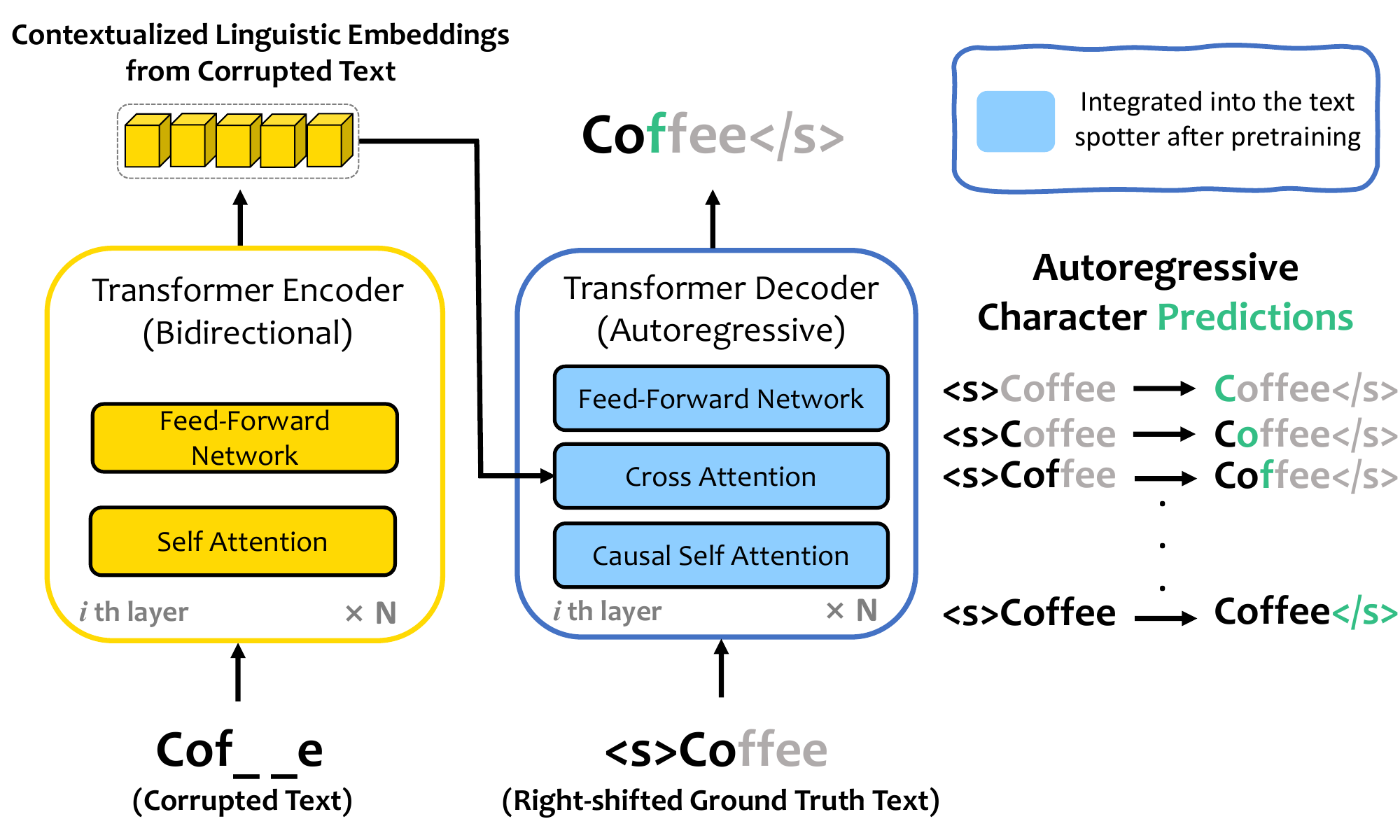}
  \caption{Proposed PLM architecture. The PLM encoder (yellow block) generates contextualized embeddings from the corrupted text, and the PLM decoder (blue block) autoregressively generates text (green) based on the encoder outputs. \model\ initializes its linguistic decoder with the PLM decoder (blue block). }
  \label{fig:plm}
\end{figure}

Figure~\ref{fig:plm} shows the architecture of our proposed language model, which consists of a transformer with a bidirectional encoder and a left-to-right decoder, inspired by BART~\cite{lewis2019bartdenoisingsequencetosequencepretraining}.
The encoder learns bidirectional context from corrupted text and the decoder predicts each token autoregressively based on its preceding tokens and the encoder output. 
The core intuition is that the decoder, trained to reconstruct words from corrupted text during the PLM pretraining, enables \model\ to better handle visually corrupted embeddings.  

Our approach differs from standard BART pretraining in three respects:
\begin{inparaenum}[i)]
\item \emph{Token granularity.}
Unlike standard BART, which uses a sub-word token vocabulary typical of PLMs, we adopt character-level tokens to better match short, fragmented scene text.
\item \emph{Input sequences.}
We pretrain the PLM on text sequences of at most three words instead of full sentences, which mirrors scene text characteristics.
\item \emph{Decoder cross attention source.}
During pretraining, the PLM decoder cross-attends to the PLM encoder states. 
However during the training and inference for \model, the decoder instead cross-attends to the projected visual features from \model's visual decoder. 
\end{inparaenum}

We follow BART’s denoising sequence-to-sequence pretraining paradigm.
Among the multiple corruption functions explored in BART (i.e., token masking, deletion, text infilling, sentence permutation, and document rotation), we adopt text infilling, masking 20-40\% of contiguous alphabetic spans while preserving non-alphabetic symbols such as digits and separators. Let \(\mathbf{y} = (y_1, \ldots, y_S)\) denote a text sequence of length \(S\), and let \(T\) be the maximum sequence length such that \(S\leq T\).  We generate a corrupted version of the sequence, \(\tilde{\mathbf{y}} = (\tilde{y}_1, \ldots, \tilde{y}_T)\), using a text infilling corruption function with padding tokens from positions \(S+1\) to \(T\).
The bidirectional transformer encoder, composed of stacked self-attention blocks, processes \(\tilde{\mathbf{y}}\) to produce contextualized hidden representations \(H^{\text{enc}} \in \mathbb{R}^{T \times D_\mathrm{L}}\), where  \(D_\mathrm{L}\) denotes language model hidden dimension size.
The left-to-right transformer decoder reconstructs the original sequence~\(\mathbf{y}\).  
Each decoder layer applies masked self-attention and cross-attention to the encoder states, followed by a position-wise feed-forward network and a linear prediction head. 

During training, we use a teacher forcing mechanism~\cite{williams1989learning} to guide autoregressive output.
The decoder receives a right-shifted target sequence \(\mathbf{y}^{\rightarrow} = (\texttt{<s>}, y_1, \ldots, y_{T-1})\) 
and predicts the next token \(y_t\) at each step based on the previous tokens \(\mathbf{y}_{<t}\) and the encoder outputs \(H_{\text{enc}}\). 
The decoder self-attention applies a causal mask \(M \in \mathbb{R}^{T \times T}\) to prevent access to future tokens:
\begin{equation}
M_{pq}= 
\begin{cases}
0,& q\le p\\
-\infty,& q>p
\end{cases}
\quad (p,q=1,\ldots,T).
\label{eq:mask}
\end{equation}
We train the model by minimizing the token-level negative log-likelihood, BART's training objective,
\begin{equation}
    \mathcal{L}_{\text{PLM}} = -\sum_{t=1}^{T}\log p\!\left(y_t \mid \mathbf{y}_{<t},\, H^{\text{enc}}\right).
    \label{eq:token-nll}
\end{equation}
At inference, the model autoregressively decodes without ground truth input, generating its own predictions until the maximum length \(T\) is reached. 

\subsection{Linguistic Decoder}\label{subsec:lingdec}

\model's linguistic decoder shares the PLM's decoder architecture.
We initialize the linguistic decoder in \model\ with the PLM's decoder weights (Section~\ref{subsec:plmtrain}), including all sub-layers in each decoder block (i.e., masked self-attention, cross-attention and position-wise feedforward network) and the linear prediction head.
This initialization transfers linguistic knowledge to \model;
although pretrained to predict text from embeddings of corrupted input, further training enables the linguistic decoder to recognize scene text by adapting to visual features.

Upon initialization, we switch the linguistic decoder’s cross-attention source from the PLM encoder states $H^{\text{enc}}$ to the projected visual features $Z^{\mathrm{vl}}$ produced by \model’s visual decoder, while keeping the architecture unchanged.
In this setup, we train the decoder on scene text data with both visual and linguistic inputs, enabling a tightly coupled integration of the two modalities in the end-to-end text spotter.
Finally, the decoder states pass through the linear prediction head to produce text recognition output autoregressively.


After integrating the linguistic decoder into \model, we use the same teacher-forcing training strategy used in PLM pretraining. 
The decoder receives the right-shifted target sequence \(\mathbf{y}^{\rightarrow}\) and attends to the visual-linguistic features \(Z^{\mathrm{vl}}\).  
We similarly train the model by minimizing the token-level negative log-likelihood:
\begin{equation}
\mathcal{L}_{\text{text}} = -\sum_{t=1}^{T}\log p\!\left(y_t \mid \mathbf{y}_{<t},\, Z^{\mathrm{vl}}\right).
\label{eq:model-lm-nll}
\end{equation}
At inference, \model\ follows the same autoregressive decoding process as in PLM pretraining, generating tokens sequentially until producing an end-of-sequence token or reaching the maximum length \(T\).






\subsection{Optimization and Loss}

Consistent with DeepSolo~\cite{ye2023deepsolo} and previous DETR-based text spotting methods \citep{zhang2022text,Huang_2023_ICCV}, we find the optimal bipartite matching between the ground truth and model outputs to calculate loss.
For both the visual encoder and visual decoder outputs, we use the Hungarian bipartite matching algorithm~\cite{kuhn1955hungarian} to pair predictions with ground truth by minimizing a cost function. 

The overall training objective of \model~is 
\begin{equation}
\mathcal{L} = \mathcal{L}_{\text{vis-enc}} + \mathcal{L}_{\text{vis-dec}} + \mathcal{L}_{\text{lang-dec}}.
\end{equation}

For the visual encoder, we use \(\mathcal{L}_{\text{vis-enc}}\), which consists of a classification loss \(\mathcal{L}_{\text{cls}}\) based on focal loss~\cite{lin2018focallossdenseobject} and a control point loss \(\mathcal{L}_{\text{coord}}\) based on \(\ell_1\) distance.
The corresponding loss weights are \(\lambda_{\text{enc-cls}}\) and \(\lambda_{\text{enc-coord}}\).
For the visual decoder, we use \(\mathcal{L}_{\text{vis-dec}}\), which includes \(\mathcal{L}_{\text{cls}}\), \(\mathcal{L}_{\text{coord}}\), and an additional boundary point loss \(\mathcal{L}_{\text{bd}}\) using \(\ell_1\) distance. 
Their respective weights are \(\lambda_{\text{dec-cls}}\), \(\lambda_{\text{dec-coord}}\), and \(\lambda_{\text{dec-bd}}\).



 
Lastly, the linguistic decoder loss \(\mathcal{L}_{\text{lang-dec}}\) incorporates the text classification loss \(\mathcal{L}_{\text{text}}\).
We use the cross-entropy loss for \(\mathcal{L}_{\text{text}}\) in an autoregressive manner, as described in Eq.~\ref{eq:model-lm-nll}, with corresponding weighting hyperparameter \(\lambda_{\text{text}}\):
\begin{equation}
\mathcal{L}_{\text{lang-dec}}
= \sum_{k}\!
\begin{aligned}[t]
  &\bigl(
      \lambda_{\text{text}} \,\mathcal{L}_{\text{text}}^{(k)}  \bigr)
\end{aligned}
\end{equation}
where \(k \in \{1, \ldots, K\}\) denotes the index of the \(k\)-th query, and \(K\) is the number of decoder queries (i.e., top encoder proposals).

\begin{table*}[!t]
\centering
\begin{adjustbox}{width=\textwidth}
\setlength{\tabcolsep}{10pt}
{\renewcommand{\arraystretch}{1}\small
\makebox[\textwidth][c]{%
\begin{tabular}{l c|cccc|cc|cc}
\toprule
\multirow{2}{*}{\textbf{Method}} & \multirow{2}{*}{\textbf{Year}} 
& \multicolumn{4}{c|}{\textbf{ICDAR 2015 }} 
& \multicolumn{2}{c|}{\textbf{Total-Text }}
& \multicolumn{2}{c}{\textbf{CTW1500 }} \\
 &  & Strong & Weak & Generic & None & None & Full & None & Full \\
\midrule
ABCNet~\cite{liu2020abcnet}                            & 2020 & 82.7 & 78.5 & 73.0 & --  & 70.4 & 78.1 & 45.2 & 74.1 \\
Mask TextSpotter v3~\cite{liao2020mask}                & 2020 & 83.3 & 78.1 & 74.2 & --  & 71.2 & 78.4 & -- & -- \\
MANGO~\cite{qiao2021mangomaskattentionguided}          & 2021 & 81.8 & 78.9 & 67.3 & --  & 72.9 & 83.6 & 58.9 & 78.7\\
ABCNet v2~\cite{liu2021abcnetv2adaptivebeziercurve}    & 2021 & 83.0 & 80.7 & 75.0 & --  & 73.5 & 80.7 & 57.5 &  77.2 \\
TESTR~\cite{zhang2022text}                             & 2022 & 85.2 & 79.4 & 73.6 & 65.3 & 73.3 & 83.9 & 56.0  & 81.5  \\
GLASS~\cite{ronen2022glass}                            & 2022 & 84.7 & 80.1 & 76.3 & --  & 79.9 & 86.2 & -- & -- \\
DeepSolo~\cite{ye2023deepsolo}                         & 2023 & 88.1 & 83.9 & 79.5 & \uline{73.8} & \uline{83.6} & \uline{89.6} & 64.2 & 81.4  \\
ESTextSpotter~\cite{Huang_2023_ICCV}                   & 2023 & 87.5 & 83.0 & 78.1 & --  & 80.8 & 87.1 & \uline{66.0}  & 83.6 \\
SPTS v2~\cite{liu2023sptsv2singlepointscene}           & 2023 & 81.2 & 74.3 & 68.0 & --  & 75.0 & 82.6 & 63.6 & \uline{84.3} \\
FastTCM~\cite{yu2023turningclipmodelscene}             & 2024 & 87.0 & 82.0 & 77.3 & --  & 79.9 & 87.2 & 60.4 & 78.8  \\
IATS~\cite{zhang2024inverselikeantagonisticscenetext}  & 2024 & 84.4 & 80.0 & 73.8 & 64.7 & 71.9 & 83.5 & 62.4  & 82.9 \\
FastTextSpotter~\cite{das2024fasttextspotter}          & 2024 & 86.6 & 81.6 & 75.4 & --  & 75.1 & 86.0 & 56.0 & 82.9 \\
LMTextspotter~\cite{xia2024lmtextspotter}              & 2024 & 87.5 & 83.4 & 78.4 & --  & 81.1 & 88.4 & 62.5 & 80.1  \\
TextBlockV2~\cite{lyu2025textblockv2}                  & 2025 & -- & -- & \uline{80.5} & --  & 80.0 & -- & 65.1 & 83.3 \\
SwinTextSpotter v2~\cite{huang2025swintextspotter}     & 2025 & \uline{89.6} & \uline{84.1} & 79.4 & --  & 82.8 & 88.4 & 61.3 & 82.0  \\
\midrule
\model\ (ours) & 2026 & \textbf{90.1} & \textbf{85.4} & \textbf{81.9} & \textbf{79.1} & \textbf{84.6} & \textbf{90.4} & \textbf{66.2} & \textbf{88.1} \\
\bottomrule
\end{tabular}}}
\end{adjustbox}
\caption{End-to-end text spotting results on ICDAR 2015, Total-Text, and SCUT-CTW1500. The standard F1-measure (Hmean) is reported.}
\label{tab:eval_overall}
\end{table*}

\section{Experiments and Results}

Here we provide a comprehensive evaluation and ablation study demonstrating how tightly coupling a pretrained language model purpose-built for scene text produces superior performance across multiple benchmarks.

\subsection{Benchmarks for Scene Text Spotting}\label{sub:scene-data}
We evaluate \model\ with \textbf{ICDAR 2015}~\cite{karatzas15icdar}, \textbf{Total-Text}~\cite{ch2017total} and \textbf{CTW1500}~\cite{liu2019curved}. 
ICDAR 2015 includes 1,000 training images and 500 testing images with word-level annotations. 
Total-Text focuses on curved and arbitrarily oriented word-level text, containing 1,255 training images and 300 testing images. 
CTW1500 differs from these datasets by emphasizing curved phrases, where each annotated unit may contain multiple words. CTW1500 includes 1,000 training images and 500 testing images.

For training, we leverage four datasets: 
\begin{inparaenum}[i)]
\item \textbf{SynthText 150K}~\cite{liu2020abcnet}, featuring 149,050 synthetic images;
\item \textbf{ICDAR 2013}~\cite{karatzas13icdar}, comprised of 229 training images and 223 testing images from the ICDAR 2013 Robust Reading competition;
\item \textbf{TextOCR}~\cite{sing21textocr}, containing 21,778 scene images for training; and 
\item \textbf{ICDAR 2017-MLT 2017}~\cite{nayaf17rrcmlt}, a multilingual scene text spotting benchmark consisting of 7,200 training images; 
we use only the Latin annotations.
\end{inparaenum}

\subsection{Datasets for PLM}\label{sub:plm-data}
To pretrain the language model for scene text, we leverage three data sources: 
\begin{inparaenum}[i)]
\item \textbf{text spotting and recognition benchmarks}: We collect the text from multiple scene text training datasets including SynthText 150K~\cite{liu2020abcnet}, Total-Text~\cite{ch2017total}, ICDAR 2013~\cite{karatzas13icdar}, ICDAR 2015~\cite{karatzas15icdar}, TextOCR~\cite{sing21textocr}, LSVT~\cite{sun2019icdar2019competitionlargescale}, ArT~\cite{chng19art}, SVT~\cite{wang2010word}, and IIIT 5K~\cite{MishraBMVC12}.
These datasets contain text in real-world scene images, typically consisting of one or two words. 
We \textit{exclude} the text in the validation and testing splits of each dataset.
\item \textbf{an NLP corpus}: We extract text from AG News~\cite{zhang2016characterlevelconvolutionalnetworkstext} and WikiText~\cite{merity2016pointersentinelmixturemodels}, which are widely used in NLP pretraining tasks.
\item \textbf{Points of Interest (POI)}: We randomly sample OpenStreetMap~\cite{OpenStreetMap} POI names in Paris, Barcelona, and Beijing, which contain POIs with Latin characters and provide diverse background and scene text styles. 
We select approximately 4,000 POIs for Paris, 4,000 for Barcelona, and 2,200 for Beijing. We only use POIs whose names consist exclusively of Latin characters. 
The entire dataset contains 932,905 unique text sequences. 
\end{inparaenum}

\subsection{Implementation and Training Details}\label{sub:implementation}
Following DeepSolo, we allow $K=100$ text proposals and sample $N=25$ points per curve. 
The visual encoder and decoder each use 6 layers with hidden size $D_\mathrm{V}=256$.
Like BART our linguistic decoder uses hidden size $D_\mathrm{L}=768$. 
The visual-to-language projection head uses a two-layer MLP to map visual to linguistic embeddings
$\mathbb{R}^{D_\mathrm{V}} \rightarrow\mathbb{R}^{D_\mathrm{L}}$. 
Loss weights are \(\lambda_{\text{coord}} = \lambda_{\text{bd}} = \lambda_{\text{cls}} = 1\) and \(\lambda_{\text{text}} = 6\) throughout the experiments, except that  \(\lambda_{\text{text}} = 15\) when fine-tuning \model\ on CTW1500. 

We pretrain \model\ on four NVIDIA A100-SXM4-40GB GPUs with a batch size of 4. For fine-tuning on each target benchmark, we train \model\ with a batch size of 2 on two GPUs (due to resource constraints).
We initialize \model's backbone, visual encoder and decoder weights from DeepSolo's pretrained weights. 
For training \model , we adopt a step learning rate schedule that starts at $10^{-4}$ and decays to $10^{-5}$ at 240K iterations and use AdamW for optimization.
We then fine-tune \model\ using the training set in the evaluation benchmarks (i.e., either ICDAR 2015, Total-Text, or CTW1500).
During fine-tuning, we adopt a step learning rate schedule that starts at $10^{-5}$ and decays to $10^{-6}$ at 45K iterations for Total-Text, and a fixed learning rate of $10^{-6}$ for ICDAR 2015. 
For CTW1500, we use a step learning rate schedule that starts at $10^{-5}$,  decays to $10^{-6}$ at 10K iterations, and further decays to $10^{-7}$ at 20K iterations. 

Compared to existing text spotting methods that predict 96 characters, our model predicts 194 tokens, including punctuation, symbols, and case-sensitive characters.
In particular, for characters that appear at the beginning of a word except for the first word in the sequence, we use distinct token variants to reflect their word-initial position, following standard practices in NLP tokenization~\cite{devlin2019bert,liu2019roberta,lewis2019bartdenoisingsequencetosequencepretraining}. For example, in the phrase “apple pie”, the \textbf{p} in “a\textbf{p}ple” and the \textbf{p} in “\textbf{p}ie” are treated as distinct tokens to reflect their word-initial positions.
The PLM matches the linguistic decoder's hidden size $D_\mathrm{L}$ across its 6 encoder layers and 6 decoder layers.
A feed-forward network (FFN) module also follows the standard Transformer architecture, which includes a dimensionality expansion from $D_\mathrm{L}=768$ to 3072, followed by a contraction back to $D_\mathrm{L}$, with GELU activation and layer normalization.
We pretrain the language model for 40 epochs with two GPUs and a batch size of 128.

\begin{figure}[t]
  \centering
\includegraphics[width=0.38\textheight,keepaspectratio]{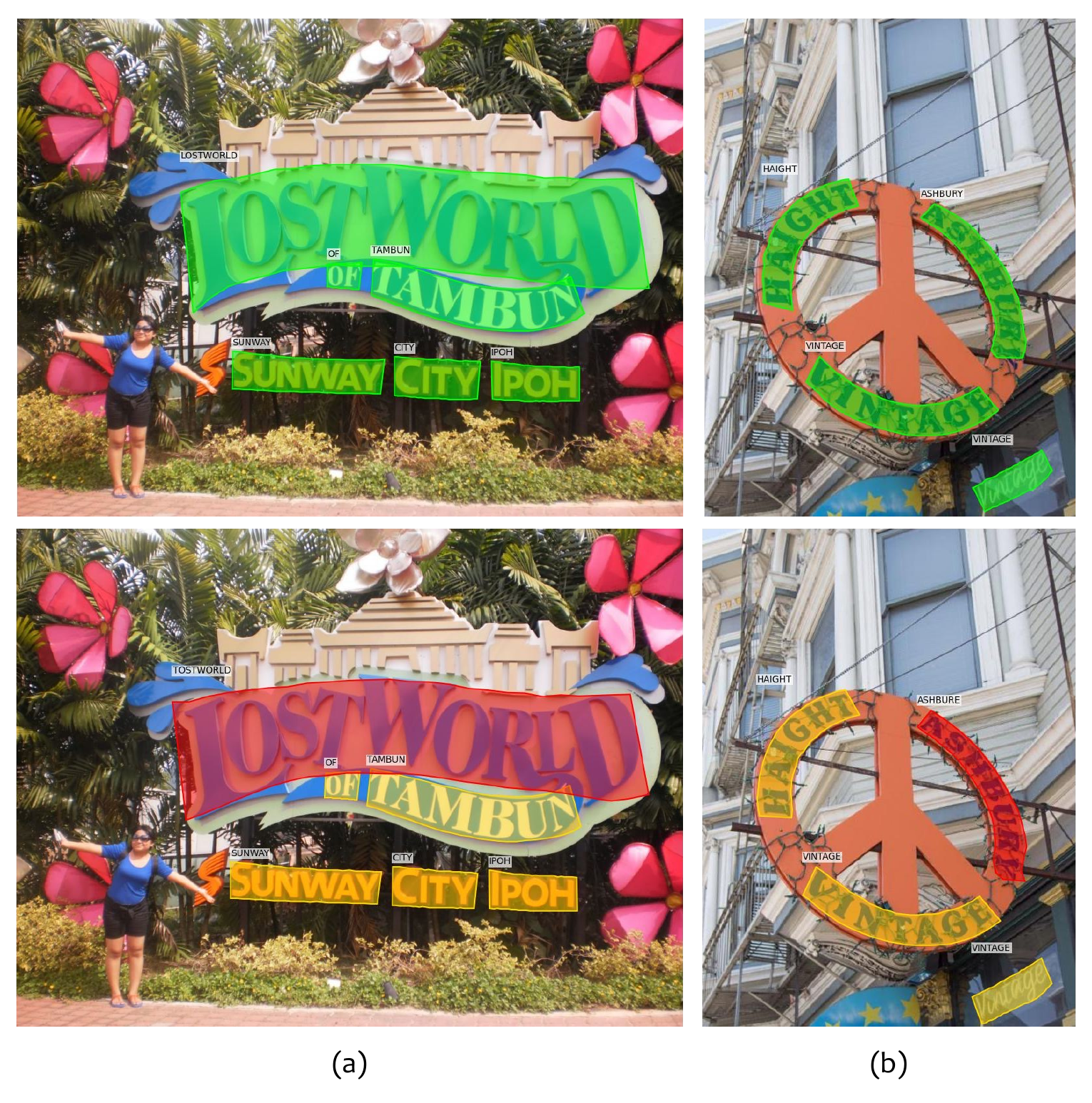}
\caption{Qualitative comparison on Total-Text (a and b) between \model\ (first row) and DeepSolo (second row). Green (ours) and yellow (DeepSolo) indicate correct detection and recognition results, while red (DeepSolo) denotes incorrect recognition results.}\label{fig:vis_compare}
\end{figure} 

\subsection{ Results and Discussion}\label{sub:experiment}
Table~\ref{tab:eval_overall} presents evaluation results on ICDAR 2015, Total-Text, and CTW1500, comparing \model\ with existing state-of-the-art end-to-end text spotting methods.
Overall, our method achieves the best performance across all datasets and standard evaluation settings. 

\paragraph{ICDAR 2015.}  
We evaluate under three standard  lexicon settings: 
\begin{inparaenum}[i)]
\item Strong (S) provides a lexicon containing the words annotated in each image.
\item Weak (W) uses the union of lexicons from all evaluation images.
\item Generic (G) contains additional words beyond the ground truth set. 
\end{inparaenum}
As a post-processing step, like TESTR and its DETR-based descendants~\cite{zhang2022text,ye2023deepsolo,Huang_2023_ICCV,huang2022swintextspotter,huang2025swintextspotter}, we use an absolute edit distance threshold of 2 when selecting a lexicon entry closest to the predicted output, discarding any detections exceeding the threshold.
\model\ surpasses all prior baseline methods, 
improving over SwinTextSpotter v2 by 0.5\% on Strong and 1.3\% on Weak, and over TextBlockV2 by 1.4\% on Generic.
In addition, \model\ achieves a lexicon-free Hmean of 79.1\%, which shows that adding the linguistic decoder boosts performance by 5.3\% over DeepSolo.

\paragraph{Total-Text.} 
We evaluate under two standard settings: without a lexicon (None) and with a lexicon constructed from all evaluation text instances (Full), applying the same edit distance-based filtering and correction.
Compared to existing methods, \model\ attains the best performance in both settings, with  improvements over 1\% on None and 0.8\% on Full compared to DeepSolo.

\paragraph{CTW1500.}  
We evaluate \model\ on CTW1500 to assess its ability to handle multi-word scene text without a lexicon (None) and with a ``lexicon'' containing the full text phrase instances from the test set.
Because of the longer text sequences in this dataset compared to others, we adopt a \emph{normalized} edit-distance threshold of 0.3 for post-processing.
\model\ achieves the best performance in both settings, yielding a 0.2\% improvement over ESTextspotter in the None setting and a 3.8\% improvement over SPTSv2 when using the lexicon. 

\paragraph{Qualitative Analysis.}

Figure~\ref{fig:vis_compare} provides a qualitative comparison between \model\ and DeepSolo, the state-of-the-art model on Total-Text. 
In Figure~\ref{fig:vis_compare}(a), our model correctly predicts the text \textit{LOSTWORLD} (9 characters), while DeepSolo mis-recognizes the first character, producing \textit{TOSTWORLD} by confusing the visually similar ``L'' and ``T''. 
Similarly, in Figure~\ref{fig:vis_compare}(b), the word \textit{ASHBURY} appears rotated against a visually complex background.
DeepSolo incorrectly recognizes the last character as ``E'', whereas \model\ accurately predicts the correct word. These examples highlight the robustness of our tightly integrated model under diverse scene text conditions. 

\subsection{Detailed Analysis}

To assess the effect of incorporating explicit linguistic information, we further analyze performance by examining word-level recognition accuracy with respect to text sequence length and vocabulary coverage, comparing our method with DeepSolo. 
Since our primary contribution lies in the recognition branch, we report accuracy only for word instances that both models detect correctly.


\sisetup{mode=match} 
\begin{table}[t]
\scriptsize 
\setlength{\tabcolsep}{2pt}
\centering
\begin{tabular}{cl S[table-format=4] S[table-format=3] S[table-format=2.1]}
\toprule
\textbf{\shortstack[c]{Word Length\\(chars)}} & \textbf{Model} & 
\textbf{\shortstack[c]{Correct Detections\\(Intersection)}} &
\textbf{\shortstack[c]{Correct\\Recognitions}} &
\textbf{\shortstack[c]{Accuracy\\(\%)}} \\
\midrule
\multirow{2}{*}{3-5}   & DeepSolo & 1055 & 933  & 88.4 \\
                       & \model   & 1055 & 953 & 90.3 \\
\cmidrule(lr){2-5}
\multirow{2}{*}{6-10}  & DeepSolo &  697 & 533  & 76.4 \\
                       & \model   &  697 & 601  & 86.2 \\
\cmidrule(lr){2-5}
\multirow{2}{*}{11+}   & DeepSolo &   39 &  17  & 43.5 \\
                       & \model   &   39 & 28   & 71.7 \\
\bottomrule
\end{tabular}
\caption{ICDAR 2015 (None) word recognition accuracy for various word lengths.}
\label{tab:acc_by_char_len_bins}
\end{table}

\paragraph{Text Sequence Length.}
We group the ground-truth text into three bins based on the number of characters in each sequence. Table~\ref{tab:acc_by_char_len_bins} shows the results for ICDAR 2015.
This factoring allows us to quantify how well each model handles short fragments compared to longer, more complex sequences.
Both models achieve relatively high accuracy on short sequences (3--5 characters), with our model slightly outperforming the baseline.
The performance gap widens for medium-length sequences (6--10 characters), where our model improves performance nearly 10\%.
The most significant difference appears for longer sequences of 11 or more characters, where our model excels by over 28\%. 
These results demonstrate that 
\model's capability of incorporating external linguistic information for scene text spotting offers significant benefits for long text sequences, particularly in the lexicon-free setting.

\sisetup{mode=match}
\begin{table}[t]
\scriptsize
\setlength{\tabcolsep}{2pt}
\centering
\begin{tabular}{l c c S[table-format=4] c c}
\toprule
\multirow{2}{*}{\textbf{Model}} & 
\multirow{2}{*}{\textbf{\shortstack[2]{PLM\\Pre-Training Text}}} & 
\multirow{2}{*}{\textbf{OOV}} & 
{\multirow{2}{*}{\textbf{Instances}}} &
\multicolumn{2}{c}{\textbf{Accuracy (\%)}} \\
\cmidrule(lr){5-6}
  & &  & & {Lexical} & {Overall} \\
\midrule
\multirow{2}{*}{DeepSolo} & \multirow{2}{*}{--}           & No & 1322 & 89.8
 & \multirow{2}{*}{84.1}\\
 &            & Yes  & 368 & 63.5
 & \\
\cmidrule(lr){3-6}
\multirow{2}{*}{\model}     & \multirow{2}{*}{Scene Text}  & No & 1322 & 93.8& \multirow{2}{*}{88.4}\\
     &   & Yes  & 368 & 68.7
 \\
\cmidrule(lr){3-6}
\multirow{2}{*}{\model}     & \multirow{2}{*}{External Text}  & No & 1466 & 92.2
 & \multirow{2}{*}{87.7}\\
     &   & Yes  & 224 & 58.0
 \\
\cmidrule(lr){3-6}
\multirow{2}{*}{\model}     & \multirow{2}{*}{External Text $\cup$ Scene Text}  & No & 
1466 & 93.3
 & \multirow{2}{*}{89.3}\\
     &   & Yes  & 224 & 62.9
 \\

\bottomrule
\end{tabular}
\caption{Comparison of word recognition accuracy between words seen during training and out-of-vocabulary (OOV) test words on ICDAR 2015 (None). 
External Text is the NLP and POI data.
}

\label{tab:vocab_inclusion_counts_acc}
\end{table}
\paragraph{Vocabulary Coverage.}\vspace{-0.5em}  
Inspired by the out-of-vocabulary (OOV) text spotting challenge~\cite{garcia-bordils22oov}, we evaluate the effect of expanding the training vocabulary. 
Table~\ref{tab:vocab_inclusion_counts_acc} summarizes the results. The first row reports the word-level accuracy of DeepSolo.
The second and third rows present the performance of \model\ initialized with PLMs trained on different corpora: one using \textit{only} text from scene text spotting datasets, and the other using \textit{only} external text data
(i.e., the NLP and POI datasets of Section~\ref{sub:plm-data}).
The last row corresponds to our proposed setting, where we train the PLM with \textit{both} scene text spotting datasets and external text.
Note that OOV counts are identical in the last two rows because fully training \model\ involves the same scene text data, regardless of how the PLM was pretrained.

Across all methods, DeepSolo and the three variants of \model\ consistently show higher accuracy on in-vocabulary words than out-of-vocabulary, regardless of the pretraining text. 
Comparing the first two rows reveals that integrating the PLM into the text spotter improves word recognition accuracy even without introducing additional vocabulary, yielding gains of 4.0\% for in-vocabulary words and 5.2\% for out-of-vocabulary words. 
Comparing the third and fourth rows further shows that learning linguistic prior knowledge from scene text during PLM pretraining is essential, as reflected by the 4.9\% boost in out-of-vocabulary performance.
Moreover, in overall accuracy, our proposed PLM configuration achieves an additional 0.9\% and 1.6\% improvement over the versions trained only on scene text and only on external text, respectively, and a net 5.2\% improvement over DeepSolo.
Finally, we observe that training \model\ without any PLM pretraining (e.g. a randomly initialized linguistic decoder) fails to converge.

\subsection{Ablation Studies}

This section presents ablation experiments to evaluate the effectiveness of a key design choice in \model: token granularity. 
We first compare an existing PLM with a subword-level token design to our character-level token design.
We then compare our token design with DeepSolo’s 96-character vocabulary, widely used in existing text spotters.
We conduct the ablation on the ICDAR 2015 test set for consistency with experiments in Section~\ref{sub:experiment}.

\begin{table}[t]
\centering
\setlength{\tabcolsep}{3pt}
\renewcommand{\arraystretch}{1.0}
{\scriptsize 
\begin{tabular}{lS[table-format=5]cccccc}
\toprule
PLM Initialization & {Vocab. size} & Granularity & S & W & G & None \\
\midrule
Our PLM (Section~\ref{subsec:plmtrain}) & 194  & character & 90.1 & 85.4 & 81.9 & 79.1 \\
BART-base            & 50265  & subword   & 83.3 & 77.2 & 73.1 & 69.0 \\
\bottomrule
\end{tabular}
}
\caption{End-to-end F1-score results (Hmean) on ICDAR 2015 comparing the linguistic decoder initializations.}
\label{tab:abl_plm_granularity}
\end{table}

\paragraph{PLM Initialization.}
To evaluate the effectiveness of our proposed PLM design for scene text spotting, we compare two initializations of the linguistic decoder: our proposed character-level PLM and a strong off-the-shelf language model.
In particular, we consider the BART decoder~\cite{lewis2019bartdenoisingsequencetosequencepretraining} pretrained on sentence-level inputs with a 50K-token subword vocabulary.
We then fully train the text spotter from each initialization, as described in Section~\ref{sub:implementation}.

Table~\ref{tab:abl_plm_granularity} shows that our PLM, with a compact vocabulary of 194 character-level tokens, consistently outperforms the BART-initialized variant.
Across different lexicon settings, our model achieves substantial gains.
The margin narrows under stronger lexicons, suggesting that lexicon guidance partially compensates for the limitations of sub-word representations by providing external linguistic constraints.
The character-level formulation could be more robust to truncated or irregular inputs common in scene text, whereas subword-based models may suffer from unstable segmentations, especially when dealing with very short phrases consisting of only one or two words.

In addition, given the 50K-token vocabulary, fine-tuning the BART-initialized variant is inherently challenging, which could also account for the weaker performance of the subword-based model.
Overall, the results demonstrate that our proposed character-level PLM provides an effective and robust initialization for scene text spotting, yielding consistent improvements over subword-based alternatives.

\paragraph{Token vocabulary.}
To isolate the contribution of our linguistic decoder, we train DeepSolo using the same 194-class vocabulary as \model.
We replace the original 96-class character classification head of DeepSolo with a 194-class head and train with the same settings used to train \model.
The results in Table~\ref{tab:vocab_size} show that even with a modified vocabulary, DeepSolo's performance falls well short of \model's, highlighting the effectiveness of the proposed linguistic decoder beyond simply enlarging the token set. 




\begin{table}[t]
\centering
{\scriptsize
\begin{tabular}{lccc}
\toprule
Methods &  Vocabulary size & None\\
\midrule
DeepSolo       & 96
& 73.8 \\
DeepSolo       & 194
& 75.5 \\
\model\ (ours)   & 194 &    79.1  &  \\
\bottomrule
\end{tabular}
}
\caption{End-to-end F1-score results (Hmean) on ICDAR 2015 comparing vocabulary size.}
\label{tab:vocab_size}
\end{table}

\section{Conclusion, Limitations, and Future Work}
We introduce \model, an end-to-end scene text spotter that tightly incorporates external linguistic knowledge from a character-level PLM. 
By explicitly leveraging pretrained linguistic knowledge and integrating it with visual features, \model\ enables robust performance across diverse and challenging cases of scene text.
Our experiments show that \model\ achieves state-of-the-art performance, demonstrating the effectiveness of tightly unified visual-linguistic modeling for scene text spotting.

Despite these performance gains, there are potential improvements to the model. Our model's parameter count is $2.73\times$ DeepSolo's (98.5M versus 36.0M); 
as a result the inference forward pass takes $1.55\times$ longer (1.35s versus 0.87s).
This increase reflects the overhead of our design, which incorporates an additional linguistic decoder. 
These observations point to opportunities for refinement.
One possible direction is to reduce redundancy by incorporating linguistic pretraining directly into the visual decoder. 
In addition, although the current inference stage in the linguistic decoder adopts greedy decoding,  incorporating a beam search could further improve accuracy, especially on long sequences, at the cost of increased computation.

\subsection*{Acknowledgment}
This material is based upon work supported in part by the National Science Foundation (Award No. 2419334) and by the University of Minnesota's Researcher Assistance Initiative for Supporting Emergencies (RAISE).

{
    \bibliographystyle{ieeenat_fullname}
    \bibliography{main,icdar24}
}

\end{document}